\documentclass{article}

\usepackage[utf8]{inputenc}
\usepackage{hyperref}
\usepackage{amsmath}
\usepackage{amssymb}
\usepackage{nicefrac}
\usepackage{geometry}
\usepackage{graphicx}
\usepackage{subfig}
\usepackage{multirow}
\usepackage{pgfplots}
\usepackage{authblk}
\DeclareUnicodeCharacter{2212}{−}
\usepgfplotslibrary{groupplots,dateplot}
\usetikzlibrary{patterns,shapes.arrows}
\pgfplotsset{compat=newest}

\usepackage{colortbl}

\newcommand{\Ha}{$\mathcal{H}$}  
\newcommand{\Hm}{\mathcal{H}}  
\newcommand{\qq}{q}  
\newcommand{\pp}{p}  
\newcommand{\mmu}{\mu}  
\newcommand{\nnu}{\nu}  
\newcommand{\ttanh}{\textsc{tanh}}
\newcommand{\relu}{\textsc{ReLU}}

\title{Pseudo-Hamiltonian Neural Networks with State-Dependent External Forces}
\author{Sølve Eidnes\thanks{Corresponding author: \texttt{solve.eidnes@sintef.no}}, Alexander J.\ Stasik, Camilla Sterud, Eivind Bøhn and Signe Riemer-Sørensen}
\affil{\small{Department of Mathematics and Cybernetics, SINTEF Digital, 0373 Oslo, Norway}}
\date{January 23, 2023}

\begin{document}
\maketitle

\begin{abstract}
\noindent  Hybrid machine learning based on Hamiltonian formulations has recently been successfully demonstrated for simple mechanical systems, both energy conserving and not energy conserving. We introduce a pseudo-Hamiltonian formulation that is a generalization of the Hamiltonian formulation via the port-Hamiltonian formulation, and show that pseudo-Hamiltonian neural network models can be used to learn external forces acting on a system. We argue that this property is particularly useful when the external forces are state dependent, in which case it is the pseudo-Hamiltonian structure that facilitates the separation of internal and external forces. Numerical results are provided for a forced and damped mass-spring system and a tank system of higher complexity, and a symmetric fourth-order integration scheme is introduced for improved training on sparse and noisy data.
\end{abstract}

\vspace{6pt}

\noindent \textbf{Keywords:} pseudo-Hamiltonian neural networks, physics informed machine learning, hybrid machine learning

\section{Introduction}\label{sec:introduction}

Hybrid machine learning is the combination of data-driven machine learning and mathematical descriptions of physical systems. The field is largely motivated by the intuition that providing physical knowledge to a learning problem will reduce the requirements for data quantity and quality. Two distinct approaches for incorporating physical knowledge in machine learning are soft and hard constraints. Soft constraints typically penalize violations of physical laws by adding penalty terms to the loss function used during training. This procedure is widely applicable, but provides no guarantees as the model must compromise between non-violation of the constraints and the predictive power as measured on available data. Hard constraints on the other hand provide mathematical guarantees of compliance with specified laws of physics. This can be achieved by enforcing the model structure, and is independent of the available data. Enforcing hard constraints introduces bias, which will typically reduce the expressiveness of the machine learning model. Hard constraints are therefore challenging to implement, as erroneous assumptions about the underlying physical system may yield wrongly biased models with poor predictive qualities.

The Hamiltonian formulation of mechanics was originally proposed in the 1830s as a generalization of classical Newtonian mechanics \cite{Hamilton1834}. Since then it has been extended and applied to mechanics, optics, electrodynamics and quantum physics, among many other fields of physics. Any closed physical system can be described by a Hamiltonian function, or the related Lagrangian. However, the Hamiltonian formulation lacks support for external interactions such as energy losses due to friction and control of the system through external forces. Such interactions are commonly present in real-world systems and crucial for engineering applications. The port-Hamiltonian formulation \cite{van2014port} has been developed to overcome these limitations, enabling interactions and energy exchanges via \textit{ports} and a corresponding Dirac structure. The port-Hamiltonian formulation has been successfully applied in various domains ranging from electrical circuits to chemistry \cite{Rashad2020twenty}.

Hamiltonian neural network (HNN) is a hybrid machine learning framework imposing hard constraints on a data-driven model \cite{Greydanus2019hamiltonian}. HNNs model the Hamiltonian function with a neural network while the system dynamics are given by the classical symplectic Hamiltonian structure. Hence, the expressiveness of the model lies in the learning of the Hamiltonian during training, while the hard-constrained structure guarantees that this learned Hamiltonian is preserved. Since the Hamiltonian framework is not designed to model non-closed physical systems, HNN models are insufficient for many practical applications. This has inspired several extensions of the HNN framework to facilitate controlled systems \cite{Zhong2020symplectic}, dissipative systems \cite{Zhong2020dissipative} and port-Hamiltonian system descriptions \cite{Desai2021port, Duong2021hamiltonian, Duong2021learning}, generalizing to situations without exact energy preservation. In this work, we consider a general \textit{pseudo-Hamiltonian} formulation that incorporates all of these extensions. This generalization of HNN makes the models more expressive and alleviates the limitations of using a hard-constrained approach.

Other extensions and improvements of HNN worth mentioning in this context are generalization to Poisson systems \cite{Jin2020learning}, generalization to coordinate-free Hamiltonian systems \cite{Chen2021neural}, and embedding the system in a higher-dimensional space and constraining to a submanifold \cite{Finzi2020simplifying, Celledoni2022learning}. However, these works only consider systems with exact energy preservation, leaving application of the proposed techniques to pseudo-Hamiltonian system descriptions to future research.

To our knowledge, the previous work most closely related to the present one is that of Duong and Atanasov \cite{Duong2021hamiltonian, Duong2021learning} where they show how to learn disturbances of a controlled system on the SE(3) manifold. However, our approach utilizes the Hamiltonian neural networks of Greydanus et al. \cite{Greydanus2019hamiltonian}, via the SymODEN \cite{Zhong2020symplectic, Zhong2020dissipative} and port-Hamiltonian neural network \cite{Desai2021port} frameworks, and is defined for systems on any manifold. Moreover, in contrast to most of the aforementioned references, we do not assume any specific structure on the Hamiltonian, like separability.


The main contributions of this paper are
\begin{itemize}
\item the introduction of pseudo-Hamiltonian neural network (PHNN) models with state-dependent external forces,
\item performance analysis of PHNN models for systems with strictly time-dependent external forces,
\item the proposal of using PHNN models to learn state-dependent external forces and a demonstration of how models can remain accurate when the forces are removed or replaced,
\item the introduction of a symmetric fourth-order integrator for accurate training with unknown derivatives without assuming any structure on the Hamiltonian,
\item the Python package \texttt{phlearn}, which has been developed and used to generate the numerical results. Documentation and standard examples can be found at \url{https://github.com/SINTEF/pseudo-hamiltonian-neural-networks}.
\end{itemize}

The rest of the paper is organized as follows: First, in Section \ref{sec:background} we provide the necessary background on Hamiltonian formulations of dynamical systems, and define the pseudo-Hamiltonian formulation that our models are based on. Then, in Section \ref{sec:phnn_time}, we present results of PHNN applied to a mass-spring system with damping and external forces, similar to the problems studied in \cite{Desai2021port}. We demonstrate that the improved performance over the most basic baseline models comes from separating the model in a state-dependent and a time-dependent part rather than from imposing a Hamiltonian structure. The main results are found in Section \ref{sec:phnn_state}; we consider a system of tanks and pipes with potential leaks, which can be viewed as state-dependent external forces. We introduce a new fourth-order symmetric integration scheme and show that this gives improved performance. Then we demonstrate how we can learn the leakages simultaneously with the full system, and obtain a model which also applies for the system without the leakage. In the brief Section \ref{sec:control}, we give an example of how the PHNN model can be used for control, before we summarize the paper and discuss the potential for future research in Section \ref{sec:discussion}.

\section{Background and methodology} \label{sec:background}
\subsection{Hamiltonian formulation}\label{subsec:hamiltonian_formulation}
The Hamiltonian formulation describes general closed systems with energy conservation. A physical system can be described by a set of $2n$ generalized coordinates: generalized positions $\qq \in \mathbb{R}^n$ and corresponding generalized  momenta $\pp \in \mathbb{R}^n$. Note that $\qq$ and $\pp$ correspond to classical positions and momenta for simple mechanical systems. The Hamiltonian $\Hm(\qq, \pp)$ describes the total energy of the system and is connected to the dynamics via
\begin{equation}\label{eq:Hstructure}
\begin{pmatrix}
    \dot{\qq} \\
    \dot{\pp}
\end{pmatrix}
=
\begin{pmatrix}
    0 & I_n \\
    -I_n & 0
\end{pmatrix}
\begin{pmatrix}
    \frac{\partial \Hm}{\partial \qq} \\
    \frac{\partial \Hm}{\partial \pp}
\end{pmatrix},
\end{equation}
where $I_n$ is the $n$-dimensional identity matrix and $\dot{q}$ and $\dot{p}$ denotes the time derivatives of $q$ and $p$. Given \Ha{} and initial conditions $\{\qq{}_0, \pp{}_0\}$, the system is fully specified. Systems may have several invariants, and the term energy is often used interchangeably with invariant even in cases when they do not correspond to the physical energy of the system.

The HNNs of \cite{Greydanus2019hamiltonian} use a neural network $\hat{\Hm{}}_\theta$ with weights $\theta$ to approximate the Hamiltonian $\Hm{}(\qq, \pp)$ of a system. Applying the general Hamiltonian structure \eqref{eq:Hstructure} with the approximated Hamiltonian yields estimates of the time derivatives $\hat{\dot{\qq}}$, $\hat{\dot{\pp}}$. Hence, the model can be trained by minimizing the difference between the estimated and true (approximated) derivatives from the training data.

Rather than restricting the study to canonical systems \eqref{eq:Hstructure}, as is done in most of the literature on HNN, we consider a general formulation that also includes non-canonical Hamiltonian systems:
\begin{equation}\label{eq:noncan}
    \dot{x} = S(x) \nabla \Hm(x), \qquad x \in \mathbb{R}^d,
\end{equation}
for some skew-symmetric matrix $S(x) = -S(x)^T \in \mathbb{R}^{d\times d}$. Such a formulation exists for any function $\Hm : \mathbb{R}^d \rightarrow \mathbb{R}$ that is an invariant of the first-order ordinary differential equation $\dot{x} = g(x)$, i.e.\
\begin{equation*}
    \frac{d  \Hm}{d t} = \nabla  \mathcal{H}(x)^T g(x) = 0.
\end{equation*}
The matrix $S(x)$  may or may not depend on $x$, and is generally not unique if $d>2$ \cite{McLachlan1999geometric}. 

\subsection{Pseudo-Hamiltonian formulation}\label{subsec:pseudo_hamiltonian_formulation}

A generalization of \eqref{eq:noncan} that includes dissipation of $\Hm$ and external forces is the pseudo-Hamiltonian formulation given by
\begin{equation}\label{eq:noncangen}
    \dot{x} = (S(x) - R(x)) \nabla \Hm(x) + f(x,t), \qquad x \in \mathbb{R}^d,
\end{equation}
where $R(x) \in \mathbb{R}^{d\times d}$ and $x^TR(x)x \geq 0$ for all $x$. This can also be viewed as a generalization of the port-Hamiltonian systems of van der Schaft \cite{van2006port, van2014port}; in contrast to those, the pseudo-Hamiltonian formulation does not include any specific structure on $f : \mathbb{R}^d \times \mathbb{R} \rightarrow \mathbb{R}^d$, and thus we do do not consider e.g.\ the passivity-preserving property that is associated with the port-Hamiltonian formulation in control theory \cite{Beattie2019robust}. There are however several recent works on identification of strictly passive systems utilizing the port-Hamiltonian formulation \cite{Cherifi2020overview, Benner2020identification, Cherifi2022non, Morandin2022port}. Compared to PHNN, these methods are typically more data-efficient, but they are less general and require more engineering and expert knowledge to be derived. A specialization of PHNN and comparison to these methods would be an interesting future study. We also note that the general formulation \eqref{eq:noncangen} is closely connected to the General Equation for Non-Equilibrium Reversible-Irreversible Coupling (GENERIC) formalism from thermodynamics \cite{Grmela1997dynamics, Ottinger1997dynamics}, and PHNN could be extended to that setting too. This would be similar to what is done by Zhang et al.\ \cite{Zhang2021gfinns}, but they do not consider external forces.

In the following, we assume that $S$ and $R$ are independent of $x$. Letting $S$ depend on $x$ while $R=0$ and $f=0$ formulates Poisson systems. A generalization of HNN to such systems is treated in \cite{Jin2020learning}.

\subsection{Pseudo-Hamiltonian neural networks}\label{subsec:phnn}
The key innovation of PHNN is to model $\Hm{}$ and $f$ in \eqref{eq:noncangen} by separate neural networks $\hat{\Hm{}}_\theta$ and $\hat{f}_\theta$ and thus learn the internal and external energy separately.
This is similar to what is done in \cite{Desai2021port}, where the authors use the term port-Hamiltonian systems for what we call pseudo-Hamiltonian systems. However, they only consider problems where the external force $f$ is known to be strictly time dependent. The system could then be modeled without the pseudo-Hamiltonian formulation by two separate neural networks: one network that models internal dynamics and depends only on the state, and one network that models the external force and depends on time alone. We will show in Section \ref{subsec:msd_performance} that such a model performs similarly to a PHNN model. Thus, we argue that the pseudo-Hamiltonian formulation is most useful in system learning when the external forces may be state dependent and the change in energy stemming from damping and external forces cannot be immediately separated.

The separation between the terms in the formulation \eqref{eq:noncangen} is obviously not unique; we could let $H$ be constant or set $S$ and $R$ to zero and any first-order ODE could still be represented by \eqref{eq:noncangen} with all the dynamics attributed to $f$. Knowing $S$ and assuming some structure on $R$ may be necessary to learn the desired formulation, and the PHNN framework is designed to incorporate prior knowledge and assumptions by setting up the different parts of the model according to these. In the examples of this paper we aim to learn damping coefficients in $R$ by learnable parameters while simultaneously learning a neural network $\hat{\Hm{}}_\theta$ that represents the energy up to a constant and a neural network $\hat{f}_\theta$ that represents the changes in the energy that cannot be attributed to damping. In practice, our models would best be utilized in an iterative process where few assumptions are made initially, but new models are trained with added information as we learn more about the system from the models. For example, $\hat{f}_\theta$ may have output dimension $<d$ if we can assume that the external forces affect only certain states directly. An example is given in Section \ref{subsec:leaks} on how imposing new assumptions on the model may lead to faster training and more accurate results. Also we note that since the PHNN model tend to learn the simplest representation, the separation between internal dynamics and external forces can in certain cases be obtained without punishing large $\hat{f}_\theta$ by a regularization term. However, regularization is often helpful or even necessary.

The lack of uniqueness in the pseudo-Hamiltonian formulation and the general unpredictability of neural networks make it difficult to provide guarantees on the PHNN model in its most general form. This paper mainly provides a practical proof-of-concept, and a further analysis of the methodology is desired and should be done on specialized models. However, the separation of the model into parts with meaningful physical interpretations makes it easier for domain experts to get a practical understanding of the model and its behaviour, and can make the model applicable for situations different from those it was trained on.

\subsection{Implementation and hyperparameters} \label{subsec:hyperparameters1}
Following \cite{Greydanus2019hamiltonian, Chen2019symplectic, Matsubara2020deep}, we use fully connected neural networks with two hidden layers of 100 neurons each to estimate the Hamiltonian and the external force. Furthermore, we use the hyperbolic tangent (\ttanh{}) and Rectified Linear Unit (\relu{}) as activation functions for the first and second hidden layer, respectively, while \cite{Greydanus2019hamiltonian, Chen2019symplectic, Matsubara2020deep} use \ttanh{} for both. This combination was discovered to significantly improve performance when applied to dynamical systems with many dimensions and high complexity, and for consistency we use it also for the lower-dimension mass-spring problem. We relate the good performance of this set-up to the fact that the true Hamiltonians of the considered systems were linear combinations of nonlinear functions, and note that other network architectures may be preferable when modeling systems with different dynamics. The output layer has no activation. The PHNNs estimate the damping coefficient by a learnable scalar parameter.

We compare the PHNNs to baseline models that estimate the left hand side of \eqref{eq:noncangen} either by one neural network or by two networks, one state dependent and one time dependent. The baseline models have the same structure as the feedforward networks in the PHNNs, but the model with only one network has 150 hidden units in each layer instead of 100, so that the PHNN models and the baseline have a comparable number of free parameters. In all experiments, we use the Adam optimizer with batch size $32$, learning rate $10^{-3}$, and the mean squared error (MSE) as loss function \cite{2014arXiv1412.6980K}.

The models are trained on an approximation of \eqref{eq:noncangen} found by a discretization method corresponding to a numerical integration scheme, as is done in \cite{Matsubara2020deep, Jin2020sympnets, David2021symplectic}. We use the implicit midpoint method in the next section. That is, we train a model
\begin{equation}\label{eq:phnnmodel}
    \hat{g}_\theta (x,t) := (S - \hat{R}_\theta) \nabla \hat{\Hm}_\theta(x) + \hat{f}_\theta(x,t)
\end{equation}
estimating the right-hand side of \eqref{eq:noncangen} using the loss function
\begin{equation}\label{eq:loss}
    \mathcal{L} = \left\lVert \frac{x^{n+1}-x^n}{\Delta t} - \hat{g}_\theta \left(\frac{x^n+ x^{n+1}}{2}, \frac{t^n+ t^{n+1}}{2}\right) \right\rVert_2^2 + \frac{\lambda}{N} \, \left\lVert \, \hat{f}_\theta\left(\frac{x^n + x^{n+1}}{2}, \frac{t^n+ t^{n+1}}{2}\right) \, \right\rVert_1,
\end{equation}
given for one data point $x^n$. The last term is $L_1$-regularization of the external force, similar to what is suggested in \cite{Desai2021port}, with $\lambda$ being the regularization parameter weighting the penalty. In Section \ref{sec:phnn_state} we introduce a new fourth-order integrator to replace the implicit midpoint method in \eqref{eq:loss}.

\section{PHNN for systems with time-dependent external forces}\label{sec:phnn_time}
The aim of this section is to evaluate the utility of using PHNNs to model systems where the external forces are strictly time dependent, whether this is prior knowledge or not. We base our study around an example similar to that studied in \cite{Desai2021port}.

\subsection{Damped and forced mass-spring system} \label{subsec:msd_system}
Consider a mass-spring system with damping, affected by an external force $f$,
\begin{equation}\label{eq:dfmassspring}
    m \ddot{x} + c \dot{x} + k x = f(x, t),
\end{equation}
where $m$ is the mass, $c$ is the damping coefficient and $k$ is the stiffness coefficient. Letting $\qq = x$ and $\pp = m \dot{x}$, such that $\qq$ is position and $\pp$ is momentum, the pseudo-Hamiltonian formulation of the system is
\begin{align}\label{eq:phdfmassspring}
\begin{pmatrix}
    \dot{\qq} \\
    \dot{\pp}
\end{pmatrix}
&=
\left[\begin{pmatrix}
    0 & 1 \\
    -1 & 0
\end{pmatrix}
-\begin{pmatrix}
    0 & 0 \\
    0 & c
\end{pmatrix}\right]
\begin{pmatrix}
    \frac{d \Hm}{d \qq} \\
    \frac{d \Hm}{d \pp}
\end{pmatrix}
+
\begin{pmatrix}
    0 \\
    f(q,p,t)
\end{pmatrix}
\end{align}
for $\Hm{}(\qq,\pp) = \frac{1}{2} k \qq^2 + \frac{1}{2m} \pp^2$. This is on the form \eqref{eq:noncangen} with $x:=(q,p)$.

Throughout the paper, we consider a forced and damped mass-spring system \eqref{eq:dfmassspring} with damping coefficient $c=0.3$ and force term $f(t) = \sin (3t)$. Initial conditions for this system are uniformly sampled, satisfying $q_0^2 + p_0^2 = r_0^2$ with $1 \leq r_0 \leq 4.5$.

\subsection{Performance analysis} \label{subsec:msd_performance}
In the following numerical experiments, we have tested four different models: two baseline models and two PHNNs. The first baseline model consists of a single neural network taking both state variables and time as input. This is what is used as a baseline model in \cite{Desai2021port}, but we also include a baseline model consisting of one strictly state-dependent neural network and one strictly time-dependent neural network, which we argue is a fairer comparison to the PHNN model presented in that paper. The PHNNs are informed that both damping and the external force only directly affect $\dot{\pp}$. That is, the damping is estimated by a single learnable parameter modelling $c$ and the external force by a single-output network modelling $f$ in \eqref{eq:phdfmassspring}. One PHNN does not assume a state-independent external force, and thus estimates $f(t)$ using a neural network $\hat{f}_\theta(q, p, t)$, while the other PHNN (correctly) assumes time-dependence only and uses a neural network $\hat{f}_\theta(t)$.

\begin{figure}[ht!]
    \centering
    \includegraphics[scale=0.99]{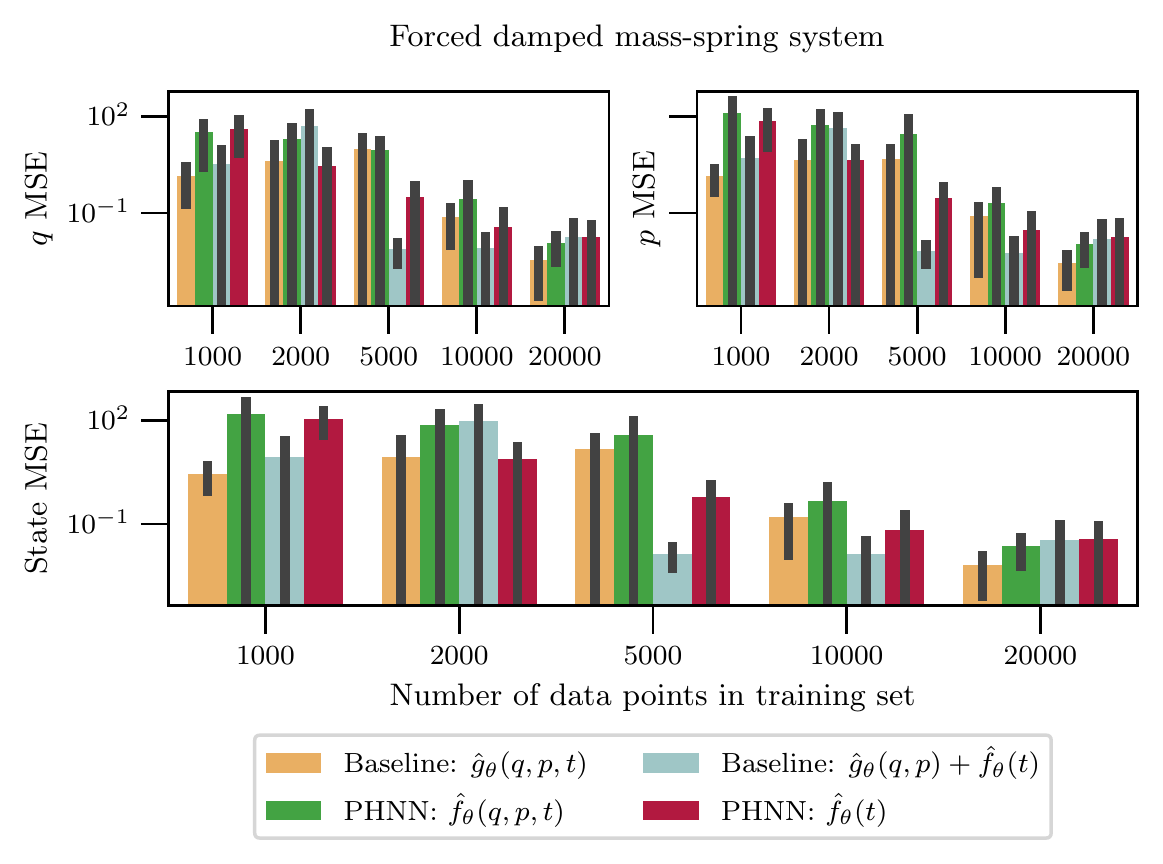}
    \caption{Mean and standard deviation of MSE of state estimates for increasing amounts of training data for the forced and damped mass-spring system.}
\label{fig:msd_datapoints}
\end{figure}

\begin{figure}[ht!]
    \centering
    \includegraphics[scale=1.]{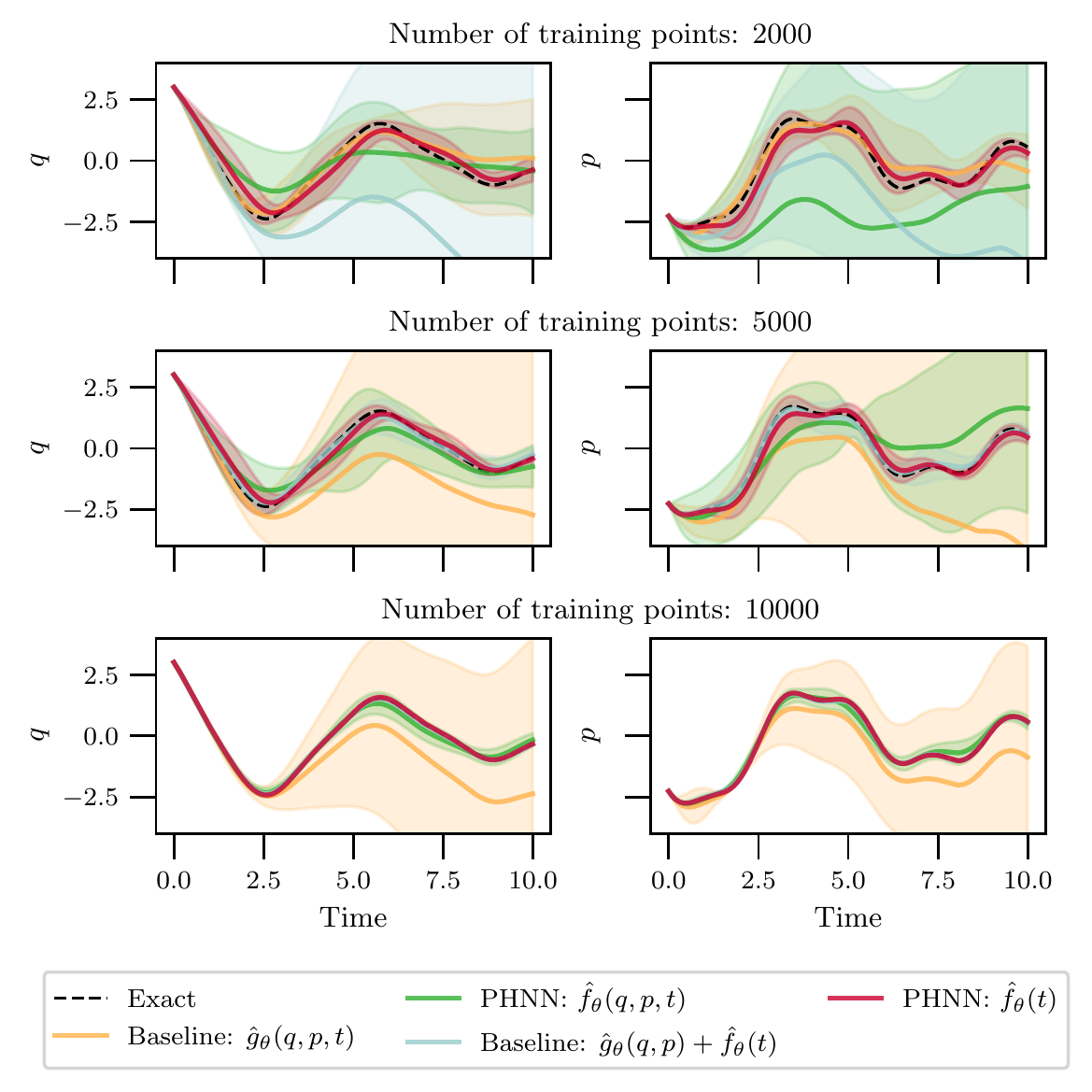}
    \caption{Example trajectory with the mean of the predictions made by the models trained on $2000$, $5000$ and $10000$ data points for the forced and damped mass-spring system. Error bands indicate standard deviation of predictions made by the 10 models of each type.}
\label{fig:msd_trajectory}
\end{figure}

We generate five data sets of $1000$, $2000$, $5000$, $10000$ and $20000$ samples. The trajectories in the data sets are all of length $10$ with sampling time $1/100$. We train the models for $20000$ epochs, with $\lambda=0.1$ for the PHNN models. Since the models rely on a non-convex optimization problem, we randomly initialize and train 10 models of each model type for each data set and also report the standard deviation as a measure of how reliable the model is if only trained once; see Figure \ref{fig:msd_datapoints}. Generally, the MSE decreases as the number of training data samples increases. Given enough training data all models learn to estimate the states well, and none of the models perform well when trained on a very limited amount of data. For medium amounts of data, the one-network baseline model and the PHNN model with state-dependent external force perform similarly, but are outperformed by the PHNN with state-independent external force and especially the two-network baseline model. Thus we conclude that, for this example, the information that contributes the most to increased performance is the separation into a state-dependent and a time-dependent term, rather than the pseudo-Hamiltonian structure. Figure \ref{fig:msd_trajectory} shows an example trajectory not in the training set, where the solid lines indicate the mean prediction made by the 10 models of each type, and the shaded areas indicate the standard deviation of the predictions.

\begin{figure}[ht!]
    \centering
    \includegraphics[scale=1.]{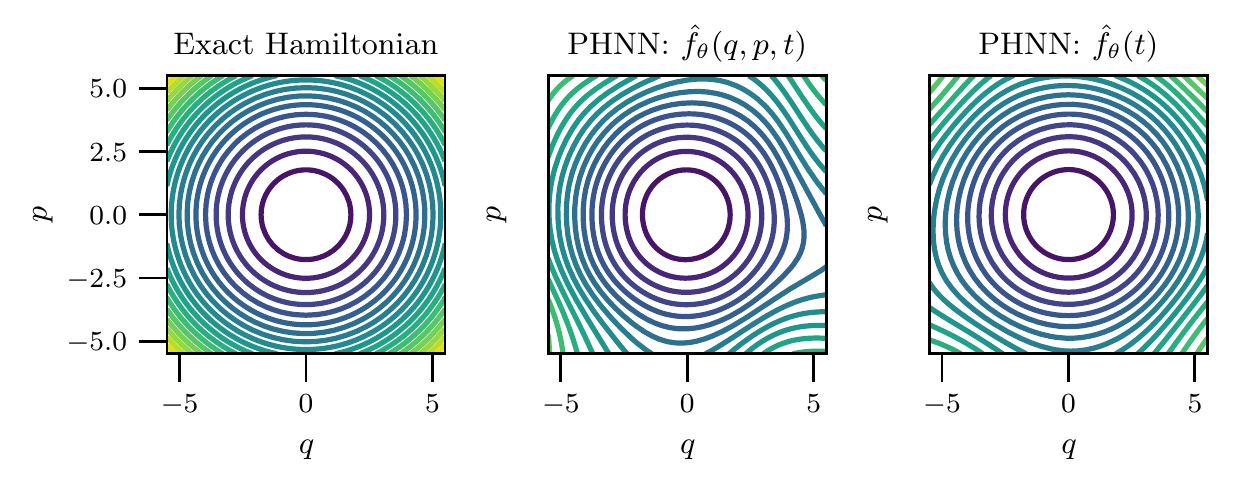}
    \caption{Contour plot of the exact Hamiltonian of the forced and damped mass-spring system along with the adjusted Hamiltonians estimated by a PHNNs with state-dependent and a PHNN with state-independent external force network.}
\label{fig:msd_hamiltonian}
\end{figure}

The learned Hamiltonian estimate will typically be offset by a constant bias, which does not affect the gradient and thus not the trajectories produced by the model either. Therefore we choose to compare the adjusted Hamiltonian $\Hm_\theta(q,p) - \Hm_\theta(0,0)$ to $\Hm(q,p)$. Figure \ref{fig:msd_hamiltonian} shows the exact Hamiltonian of the forced and damped mass-spring system along with the adjusted Hamiltonians estimated by the two PHNNs. The model of each type with the lowest MSE when estimating the gradient of the Hamiltonian was chosen. Note that due to the initial condition sampling scheme, $-4.5 \leq p, q \leq 4.5$ in the training data, and outside this area the prediction of the Hamiltonian quickly deteriorates.

\begin{figure}[ht!]
    \centering
    \includegraphics[scale=1.]{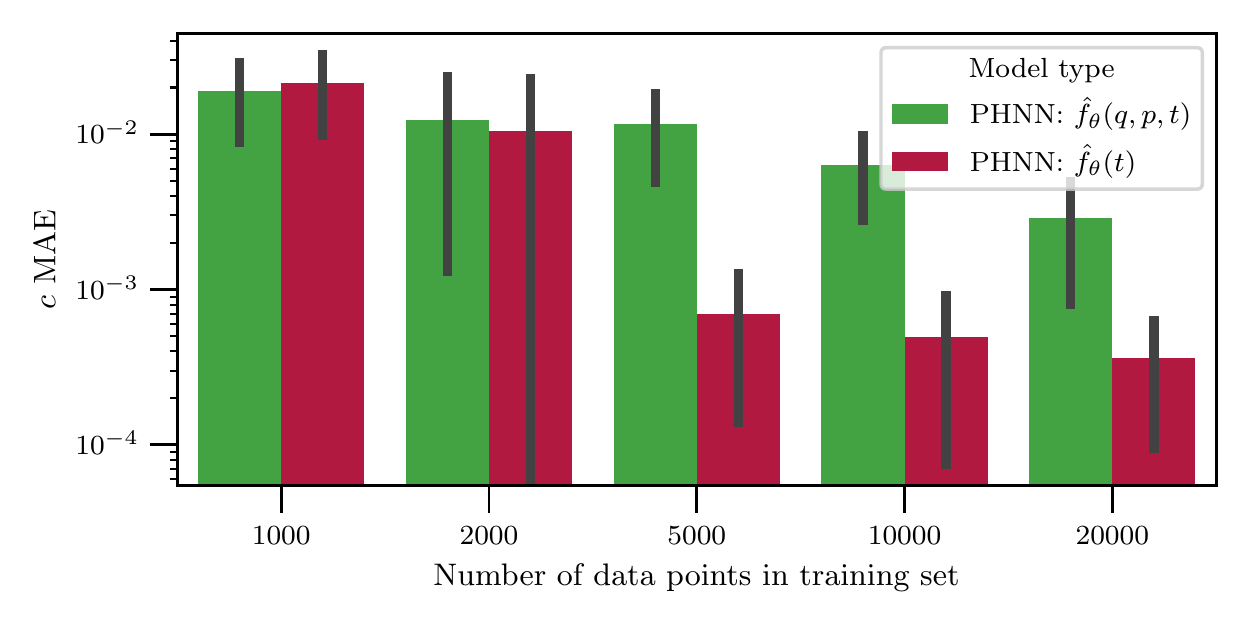}
    \caption{Mean and standard deviation of the mean absolute error in the estimates of the damping coefficient $c$ for the forced and damped mass-spring system.}
\label{fig:msd_damping}
\end{figure}

\begin{figure}[ht!]
    \centering
    \includegraphics[scale=1.]{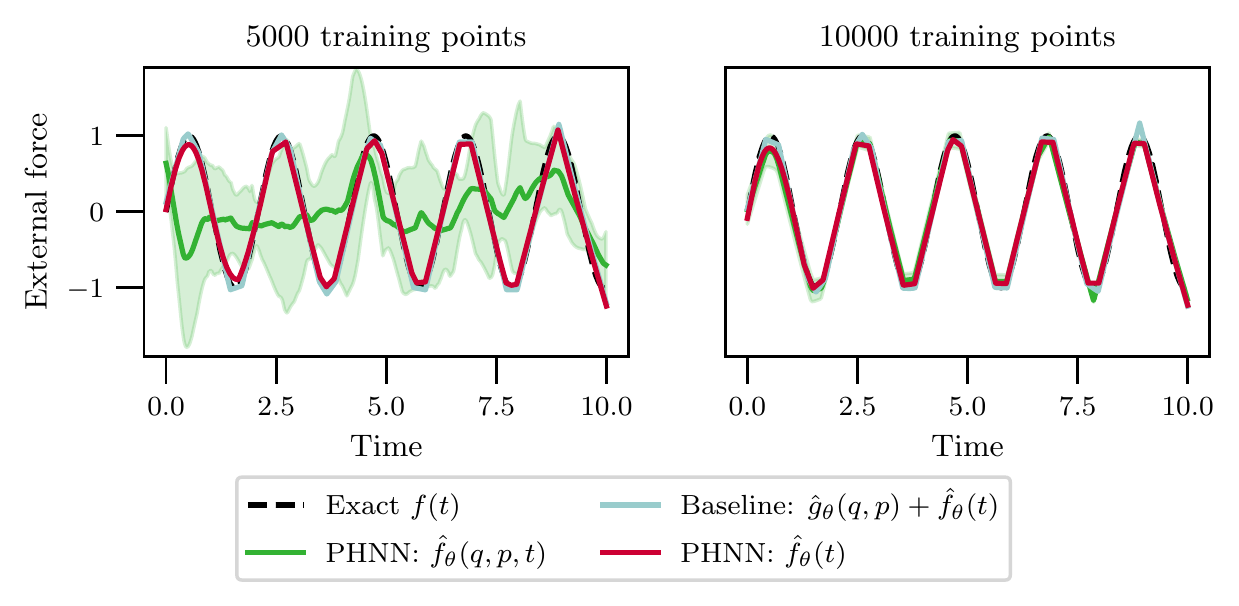}
    \caption{External force estimated by the two-network baseline model and the two PHNN model types. The error band is included to show the standard deviation of the estimate made at each time by the PHNN with a state-dependent external force network. At each time $T$, the mean and standard deviation of $\hat{f}_\theta(q(T), p(T), T)$ is computed over all values of $q(T), p(T)$ occurring in the test set.}
\label{fig:msd_external_force}
\end{figure}

\begin{figure}[ht!]
    \centering
    \includegraphics[scale=1.]{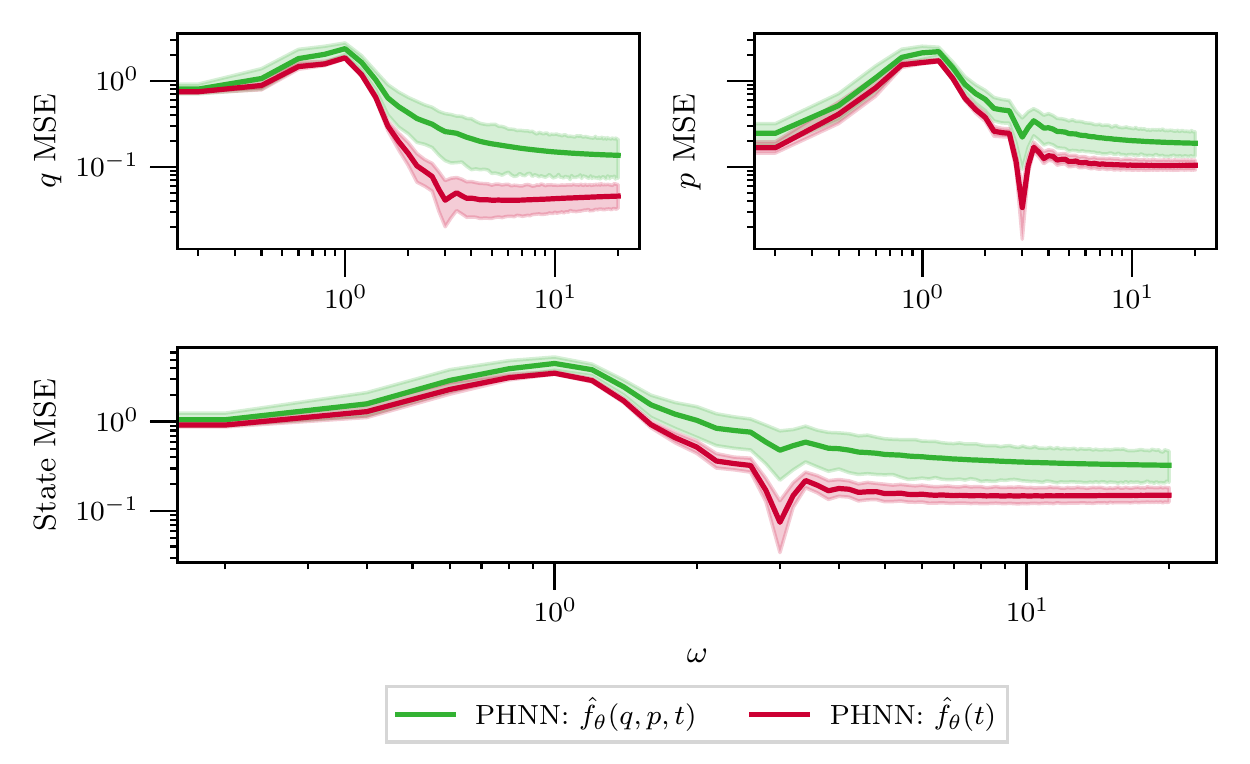}
    \caption{Mean and standard deviation of the MSE of state estimates when the learned external force in the model is replaced with a control input $f(t)=\sin{(\omega t)}$, for different $\omega$. The models are trained on the data set with 10000 samples.}
\label{fig:msd_diff_freq}
\end{figure}

Figure \ref{fig:msd_damping} compares the mean absolute error of the learned damping coefficient for the two types of PHNNs as a function of training data sample size. Figure \ref{fig:msd_external_force} shows how well the PHNNs estimate the external force. Since the models can only learn the separation of the external force from the internal system up to a constant, we subtract the time-average of the external force from the learned external force before comparing to the exact solution. Here, we have chosen the PHNN of each type with the lowest MSE on estimating the external force during testing. In Figure \ref{fig:msd_diff_freq} we show how the model adapts if the learned external force is replaced by a known input of different frequencies. The error is smallest for $\omega=3$, which corresponds to the external force in the training data, and it handles higher frequencies better than lower.

\section{PHNN for systems with state-dependent external forces}\label{sec:phnn_state}
Since we argue that the main use of PHNN is to learn external forces where these cannot be assumed to be strictly time dependent, we present here an example where the system under consideration has state-dependent external forces.

\subsection{Connected tanks in a pseudo-Hamiltonian formulation} \label{subsec:tank_system}

We construct a system of $N$ tanks connected by $M$ pipes, as described in \cite{Schaft2013graphs}. The system can be regarded as a directed graph $\mathcal{G} = (\mathcal{V, E})$ with the vertices $\mathcal{V}$ representing the tanks and the edges $\mathcal{E}$ representing the pipes. With $\nnu_i$ being the flow through pipe $i$, $\mmu_j$ the volume of the fluid stored in tank $j$, and $B$ the incidence matrix of the graph, conservation of volume gives
\begin{equation}\label{eq:tank1}
    \dot{\mmu} = -B\nnu.
\end{equation}
We assume that the flow through pipe $i$ is given by
\begin{equation}\label{eq:tank2}
    J_i \dot{\nnu{}}_i = P_k - P_l - \lambda_i(\nnu{}_i),
\end{equation}
where $J_i$ depends on the density of the fluid and the pipe dimension, $P_k$ and $P_l$ are the pressures in either end of the pipe, and $\lambda_i(\nnu_i)$ is the friction term \cite{dePersis2011pressure}. In our numerical experiments we use $\lambda_i(\nnu_i) = r_i \nnu_i$ and hence assume that the $R$ from \eqref{eq:noncangen} is independent of the state variables. Generalization to account for a more expressive friction term is entirely feasible within the framework, but not considered here. The kinetic energy stored in the flow within a pipe is given by $E_i^\text{pipe} = \frac{1}{2}J_i\nnu_i^2$. The gravitational pressure on the bottom of a tank $j$ is given by $P_j = \rho g \frac{\mmu_j}{A_j}$ with $A_j$ as the tank footprint, $\rho$ as the density of the fluid and $g$ as the gravitational constant, and the associated potential energy in the tank is given by $E_j^\text{tank} = \frac{g \rho}{2 A_j} \mmu_j^2$. Substituting the flow with the energy variable $\phi_i := J_i \nnu_i$, we get the following Hamiltonian:
\begin{equation}
    \Hm(\phi, \mmu) = \sum_i^M \frac{1}{2 J_i}\phi_i^2 + \sum_j^N \frac{g \rho}{2 A_j} \mmu_j^2.
\end{equation}
In the absence of friction or external forces, we can rewrite \eqref{eq:tank1} and \eqref{eq:tank2} as \eqref{eq:noncan} with $x:=(\phi, \mmu)$ and
\begin{equation}
    S = \begin{bmatrix}
0_{M\times M} & B^T\\
-B & 0_{N\times N}
\end{bmatrix}.
\end{equation}
Including friction and external forces, we get the pseudo-Hamiltonian system
\begin{equation} \label{eqn:tanks}
\begin{bmatrix}
\dot{\phi}\\
\dot{\mmu}
\end{bmatrix} = \begin{bmatrix}
-R_p & B^T\\
-B & 0_{N\times N}
\end{bmatrix}
\begin{bmatrix}
\frac{\partial \Hm}{\partial \phi}\\
\frac{\partial \Hm}{\partial \mmu}
\end{bmatrix} + 
\begin{bmatrix}
f_p(\phi, \mmu, t)\\
f_t(\phi, \mmu, t)
\end{bmatrix},
\end{equation}
where $R_p$ is an $M \times M$ diagonal matrix with elements $r_i$, and $f_p$ and $f_t$ are the external forces acting on the pipes and the tanks, respectively.

\begin{figure}[ht]
    \centering
    \includegraphics[width=0.3\textwidth]{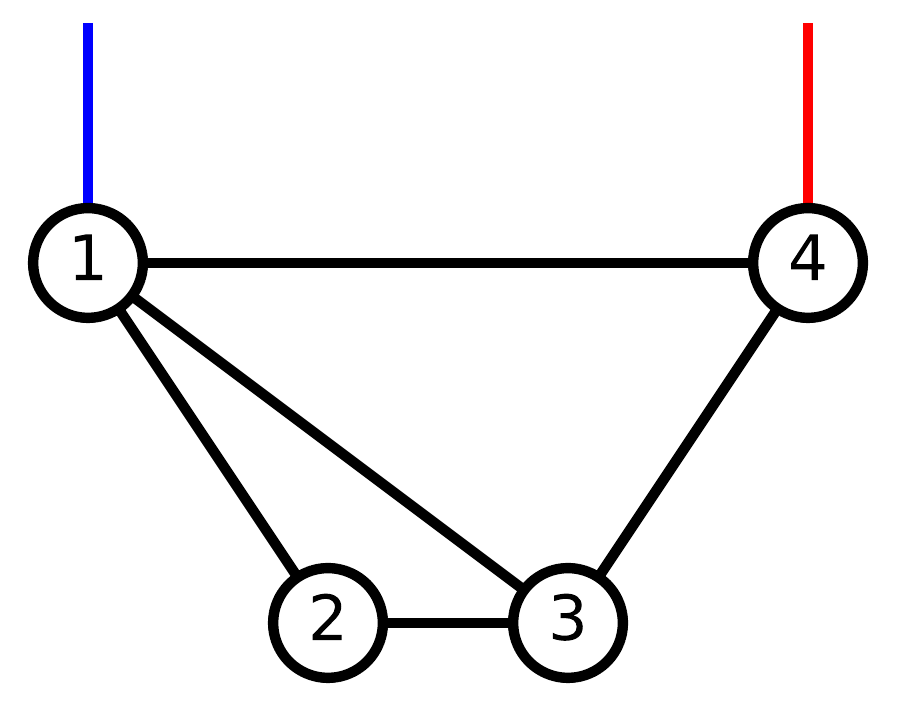}
    \caption{Graph showing how the five pipes connect the four tanks (1-4) and the external force on the fourth tank (red).}
\label{fig:graph_tank_system}
\end{figure}

Throughout this section, we consider a tank system with four tanks and five pipes, connected as shown by the graph in Figure \ref{fig:graph_tank_system}. Unless otherwise specified, we use the following parameters when simulating the system: $\rho=1$, $J_i=0.02$ for all $i$, $A_j = 1$ for all $j$, $R_p = \mathrm{diag}(0.03, 0.03, 0.09, 0.03, 0.03)$. For convenience, no units are specified and the problem is considered as scale-free. An external, state-dependent force acts on the fourth tank, so that $f_p = (0,0,0,0,0)$ and $f_t = (0,0,0, -10 \, \mathrm{min} (0.3, \, \mathrm{max}(\mmu_4, -0.3)))$. Except for the experiments in Section \ref{subsec:leaks}, the fact that only the fourth tank is affected by an external force is assumed to be known. Initial conditions are uniformly sampled such that $-1 \leq x_0 \leq 1$ for all states.

\subsection{Choice of discretization method in training} \label{subsec:discretization}
Instead of training on the integration scheme as we do, works like \cite{Greydanus2019hamiltonian,Chen2019symplectic, Desai2021port} either assume that derivatives of the state variables are known or perform one or more integration steps at each training step. Having the exact derivatives is unlikely in real-world scenarios, and thus we do not use those in our experiments. Performing integration in the training necessitates the use of explicit integrators or requires a large increase in computational cost. Specifically, it prohibits efficient use of symmetric integrators like the implicit midpoint method. In early testing, we observed that using symmetric integration schemes in the training greatly improved the performance of our tank system models, which motivated the development of a symmetric fourth-order integrator specifically designed for the inverse problem of learning a system from data, rather than integration.

For reasons of brevity, we consider autonomous systems $\dot{x} = g(x)$ in this section. We define a general integrator
\begin{equation*}
    \frac{x^{n+1}-x^n}{\Delta t} = \Phi_{\Delta t} (g, x^n, x^{n+1})
\end{equation*}
so that $\Phi_{\Delta t} (g, x^n, x^{n+1}) = g((x^n+x^{n+1})/2)$ gives the implicit midpoint method resulting in \eqref{eq:loss}.  Different integrators have been proposed in the literature on HNN, building on established theory from the field of numerical integration, e.g.\ on symplectic methods for systems with invariants \cite{Hairer06}. However, the inverse problem of learning an ODE from data is different from integrating a known system, which alters which properties of numerical integrators we will want to consider. For instance, a certain class of implicit integrators called mono-implicit Runge--Kutta (MIRK) methods \cite{Bokhoven1980efficient, Cash1082mono} do not depend on intermediate steps and are thus explicitly given by the known data in the inverse problem. Moreover, these integrators may be less expensive than comparable explicit integrators when used for training.

To obtain a more accurate discretization of the system than obtained by the second-order implicit midpoint method, we propose the fourth-order scheme
\begin{equation}\label{eq:srk4}
\begin{split}
\frac{x^{n+1}-x^n}{\Delta t}= & \, \frac{1}{2} g\left(\frac{x^n+x^{n+1}}{2}-\frac{\sqrt{3}}{6} \Delta t \, g\Big( \big(\frac{1}{2}+\frac{\sqrt{3}}{6}\big)x^n + \big(\frac{1}{2}-\frac{\sqrt{3}}{6}\big)x^{n+1}\Big)\right)\\
& + \frac{1}{2} g\left( \frac{x^n+x^{n+1}}{2}+\frac{\sqrt{3}}{6} \Delta t \, g\Big( \big(\frac{1}{2}-\frac{\sqrt{3}}{6}\big)x^n + \big(\frac{1}{2}+\frac{\sqrt{3}}{6}\big)x^{n+1}\Big)\right),
\end{split}
\end{equation}
which is symmetric but not symplectic. As a numerical integrator it is an implicit Runge--Kutta method, and more specifically a MIRK method, but as a discretization of \eqref{eq:noncangen} it is explicitly given by $x^n$ and $x^{n+1}$. This distinguishes it from e.g.\ the Gauss--Legendre method of order four, which would require a system of equations to be solved by e.g.\ Newton's method at each training step. Moreover, the implicit midpoint method and \eqref{eq:srk4} are applicable for non-separable Hamiltonian systems, in contrast to the second-order leapfrog method used in \cite{Chen2019symplectic} and Yoshida's fourth-order method, used in \cite{DiPietro2020sparse, Desai2021variational}.

The computational cost of a method used during training is dominated by the number of evaluations of $g$. Thus, the implicit midpoint method is comparable to the forward Euler method, while \eqref{eq:srk4} is approximately four times as expensive and comparable to the classic Runge--Kutta method. The advantage of using a higher-order method is generally most prevalent when data is sparse, while symmetric methods deal well with noise. Thus, even though \eqref{eq:srk4} generally performs best, the implicit midpoint method might be preferable when the training data is obtained at a sufficiently high frequency and computational cost is an issue.

Further analysis of the scheme \eqref{eq:srk4} and more high-order integration schemes is outside the scope of this study and will be the topic of a paper in preparation, which considers the general challenge of applying any data-driven model for estimating a dynamical system described by differential equations.

\subsection{Comparison of discretization methods for the tank system}\label{subsec:discretization_comp}
Consider the system of four tanks and five pipes described in Section \ref{subsec:tank_system}. Four different discretization methods are used for training PHNNs modeling the tank system for six different data sets. The four discretization methods are the forward Euler method, the classic Runge--Kutta method (RK4), the implicit midpoint method, and the symmetric fourth-order scheme \eqref{eq:srk4} (SRK4). The six data sets consist of trajectories lasting one time unit and are made to reflect different levels of data quality:
\begin{itemize}
    \item low sampling time ($1/100$) and many samples ($30000$) without noise
    \item low sampling time and many samples with moderate noise (Gaussian noise with a standard deviation $\sigma = 0.03$ added to the measurements of the states)
    \item low sampling time and many samples with much noise (standard deviation $\sigma = 0.05$)
    \item high sampling time ($1/30$) and few samples ($3000$) without noise
    \item high sampling time and few samples with moderate noise
    \item high sampling time and few samples with much noise
\end{itemize}

\begin{figure}[ht!]
        \centering
        \includegraphics[scale=1.]{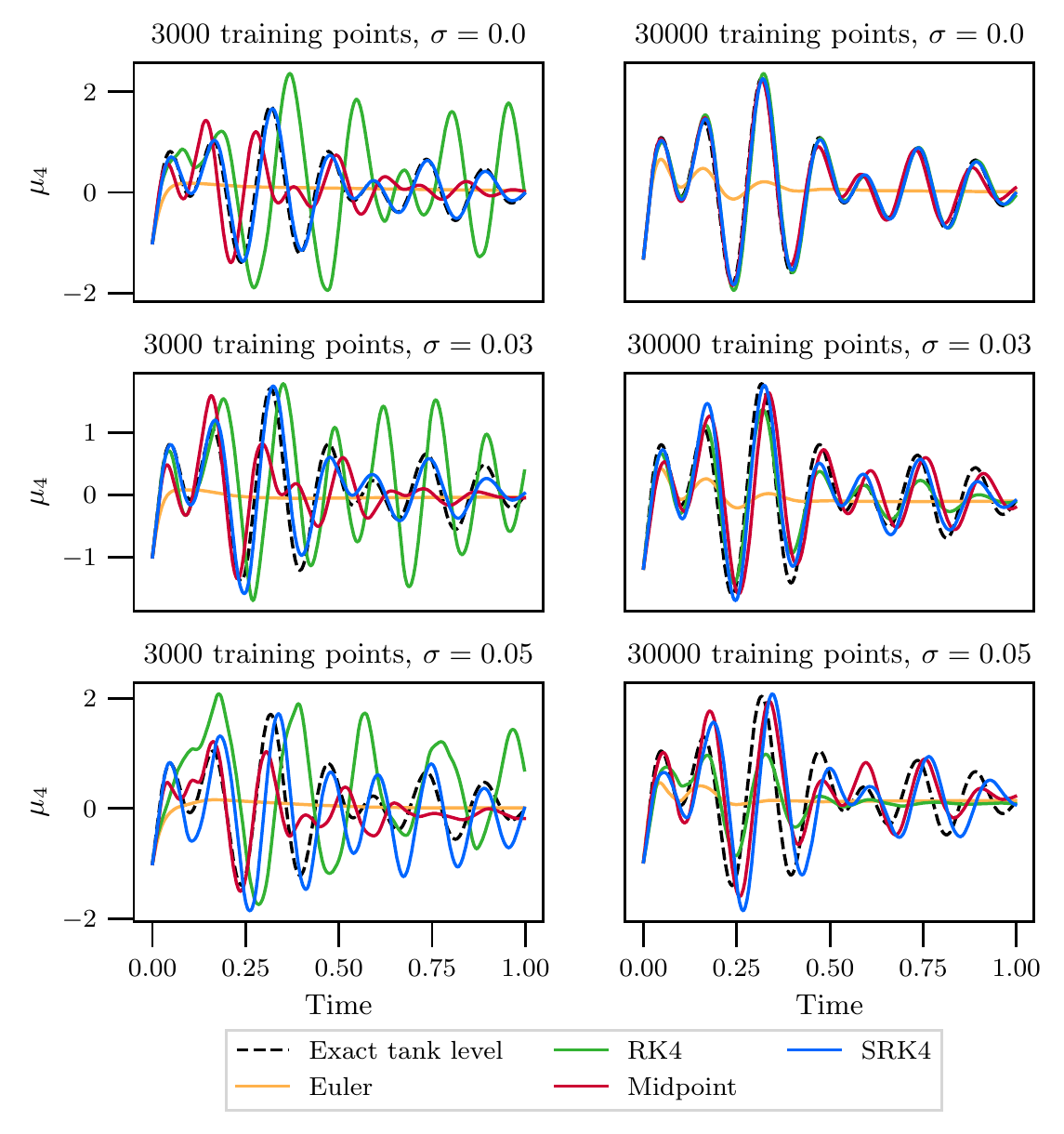}
        \caption{Volume of the fourth tank as predicted by the PHNNs with the lowest MSE on the test set. The initial condition is $\phi^0 = (-1, -1, 0, \frac{1}{2}, -1)$, $\mmu^0= (1, 1, -\frac{1}{2}, -1)$.}
\label{fig:integrators}
\end{figure}

\begin{figure}[ht!]
        \centering
        \includegraphics[scale=1.]{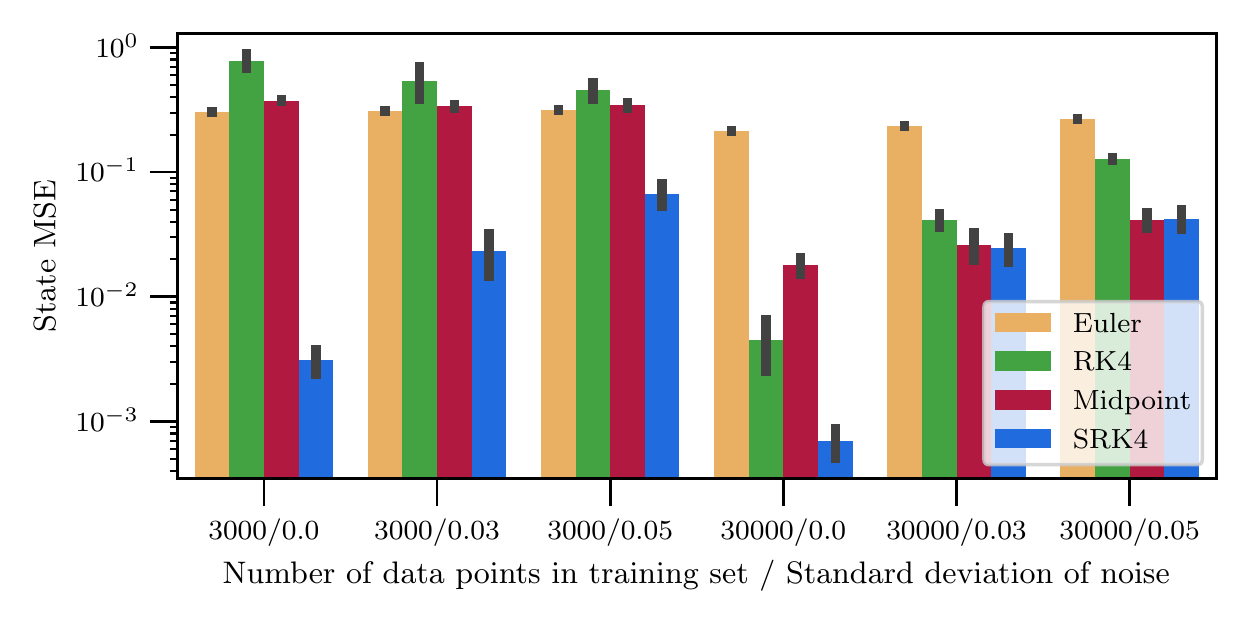}
    \caption{The mean and standard deviation of the MSE of the PHNNs trained with the different integrators on each data set. The MSE is of the predicted volume and flow in all tanks and pipes from $t=0$ to $t=1$ on the test set.}
\label{fig:integrators_barplot}
\end{figure}

For each data set and for each discretization method, 10 PHNNs are trained for 1000 epochs. A test set consisting of 10 trajectories with random initial conditions is used to test the performance of the models.
Figures \ref{fig:integrators} and \ref{fig:integrators_barplot} and Table \ref{tab:integrators_R} demonstrate that the symmetric methods handle noise well. When data is scarce a higher order symmetric method is superior to the second-order implicit midpoint method, although its advantage in approximating derivatives more accurately is less important on noisy data, where the noise becomes a limiting factor for the accuracy of the models.

\begin{table}
    \centering
    \resizebox{\textwidth}{!}{
    \begin{tabular}{l|c|c|c|c|c|c}
    \hline
    \multirow{2}{*}{} & \multicolumn{3}{ c| }{3000 training points} & \multicolumn{3}{ c }{30000 training points} \\
    \cline{2-7}
    & no noise & $\sigma=0.03$ & $\sigma=0.05$ & no noise & $\sigma=0.03$ & $\sigma=0.05$\\
    \hline
    Euler & $43.95\pm13.12$ & $44.62\pm13.16$ & $45.70\pm13.36$ & $11.38\pm3.25$ & $12.35\pm3.52$ & $13.98\pm3.94$ \\
    RK4 & $-0.59\pm0.60$ & $-0.22\pm0.50$ & $0.25\pm0.50$ & $1.08\pm0.06$ & $1.98\pm0.33$ & $3.56\pm0.81$ \\
    Midpoint & $1.59\pm0.17$ & $1.72\pm0.21$ & $1.74\pm0.25$ & $1.07\pm0.04$ & $1.21\pm0.16$ & $1.31\pm0.18$ \\
    SRK4 & $1.07\pm0.03$ & $1.21\pm0.13$ & $1.24\pm0.15$ & $1.03\pm0.02$ & $1.21\pm0.18$  & $1.19\pm0.19$ \\
    \hline
    \end{tabular}
    }
    \caption{Mean and standard deviation of the predicted friction coefficients, relative to the ground truth $R_p = (0.03, 0.03, 0.09, 0.03, 0.03)$ (i.e.\ so that $1$ would mean the correct coefficient).}
    \label{tab:integrators_R}
\end{table}

\subsection{Model performance}\label{subsec:tank_performance}

Consider the tank system described in Section \ref{subsec:tank_system}. We use this system to generate data sets with $100$, $250$, $500$, $1000$, $2500$, $5000$, $10000$ and $20000$ samples, as well as a validation set consisting of $500$ samples. The trajectories in the data sets are all of length $1$ with sampling time $1/100$ and initial conditions sampled from independent uniform distributions $\mathcal{U}(-1, 1)$.

We consider two model types: A baseline network and a PHNN. All networks take the current state of the system as input and none depend on time. The PHNN is informed that damping only directly affects the states related to pipe flow, and that the external forces are only affecting the last tank state. For each data set we train 10 models of each model type for 20000 epochs, with no regularization. For this higher-dimensional problem we used a batch size of 256, but the hyperparameters are otherwise as specified in Section \ref{subsec:hyperparameters1}.

\begin{figure}[ht!]
    \centering
    \includegraphics[scale=1.]{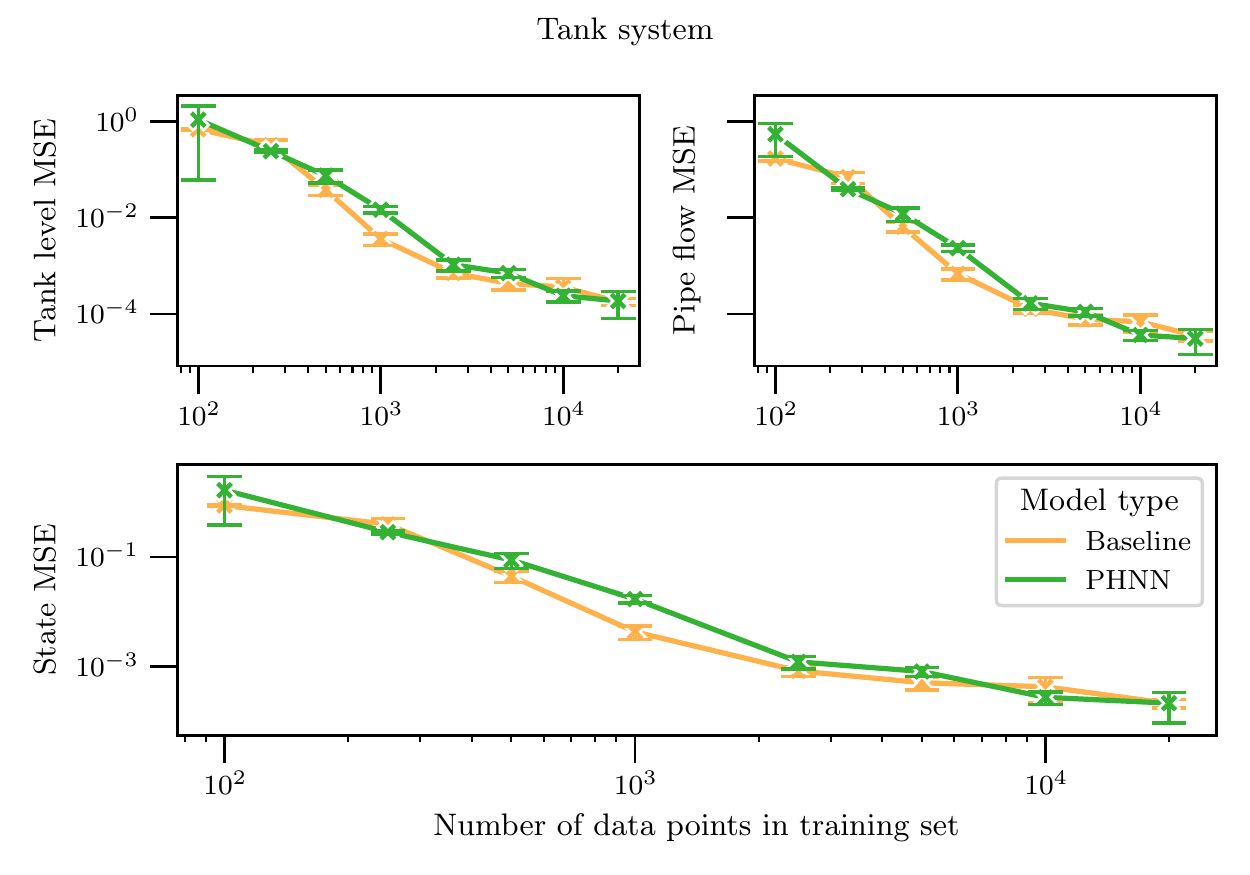}
    \caption{Mean and standard deviation of the MSE of state estimates for increasing amounts of training data for the tank system.}
\label{fig:tank_datapoints}
\end{figure}

Figure \ref{fig:tank_datapoints} shows that the two models perform comparatively and that the MSE decreases as the data size increases. As opposed to the mass-spring system in Section \ref{subsec:msd_performance}, where we could improve performance by correctly assuming that the external force was state independent, the external forces in the tank system are known to be state dependent. This leads to non-uniquely separable terms in \eqref{eqn:tanks}. However, we observed that in the case of the external force only affecting one state variable directly and this knowledge being used in the model, the imposed structure ensured that the terms were correctly separated by the model, even without using regularization. Thus the PHNN has the advantage of estimating individual terms, which we can compare to our assumptions and physical knowledge about the system.

\begin{figure}[ht!]
    \centering
    \includegraphics[width=0.98\textwidth]{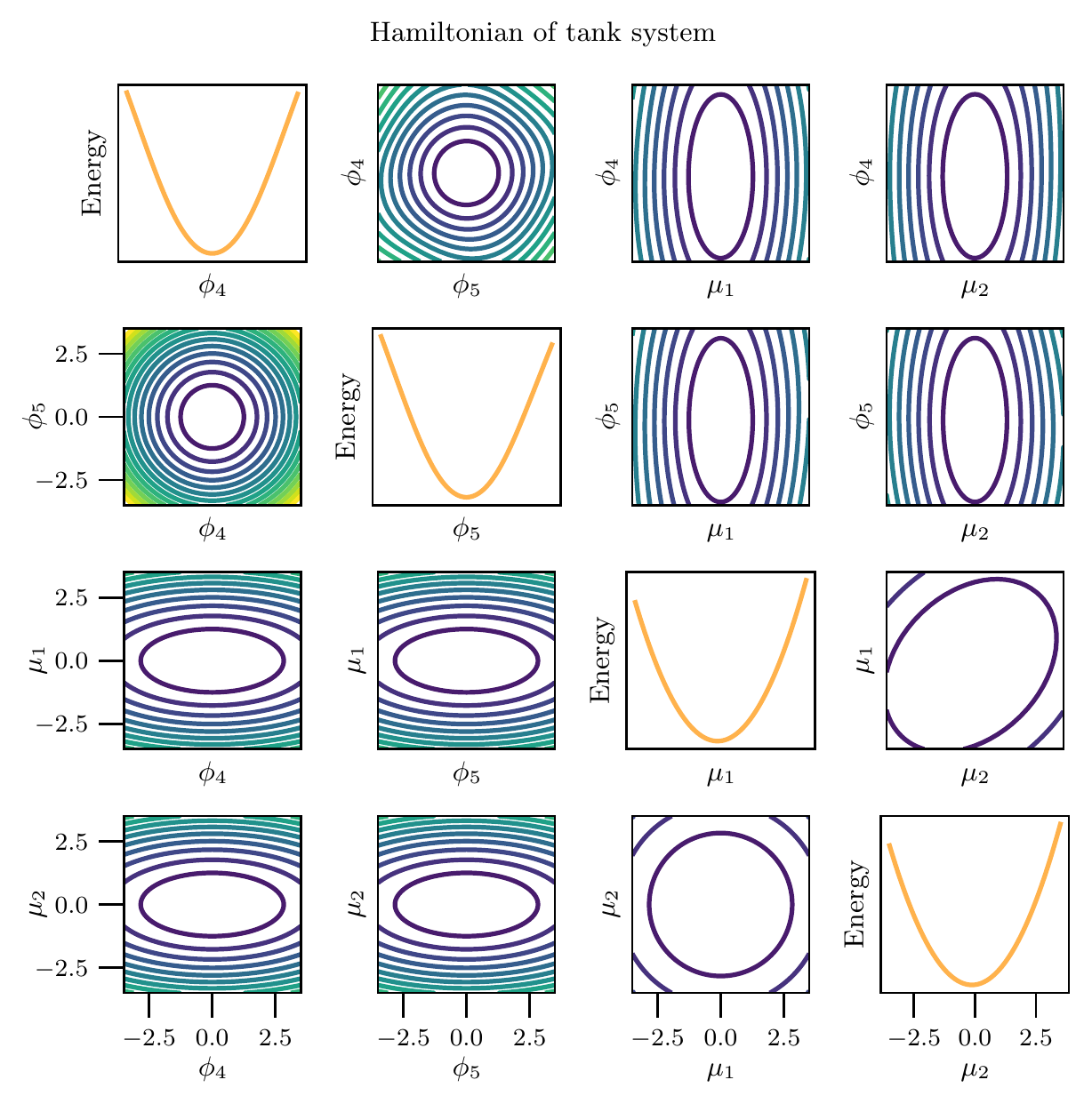}
    \caption{The lower left plots show the contour of the exact Hamiltonian when all states are set to zero except for the two displayed in each plot. The upper right shows the contour of the Hamiltonian as estimated by the PHNN with the lowest MSE in estimating $\nabla \Hm$, and adjusted by subtracting $\Hm(0,0)$. The plots on the diagonal show the estimated Hamiltonian when all states are set to zero except the one noted on the $x$-axis. }
\label{fig:tank_hamiltonian}
\end{figure}

Figure \ref{fig:tank_hamiltonian} shows the contours of the exact and estimated Hamiltonian for slices of the state space. The estimates are based on the PHNN with the lowest MSE for $\nabla \Hm$ and they are adjusted as explained in Section \ref{subsec:msd_performance}. As for the mass-spring system, and as expected, we observe that the Hamiltonian is most accurately estimated towards the center of the training data distribution. The exact Hamiltonian is spherical in the $\phi_4-\phi_5$ and $\mu_1-\mu_2$ planes, and ellipsoid in the remaining planes. The learned Hamiltonian is elongated along the same directions, but with a small offset in the $\phi_4$ and $\phi_5$ directions. The largest difference is for the $\mu_1-\mu_2$ plane, which is offset and slightly elongated. 

\subsection{Learning external forces}\label{subsec:leaks}

The PHNN framework allows for detection of unknown external forces, like a leakage or unknown inflow in one or more tanks. Moreover, in the case that external forces are altered, for instance by stopping a leak, a PHNN learned from data with a leak does not necessarily have to be retrained, as the PHNN's external forces can be altered correspondingly.

\begin{figure}[htb!]
    \centering
    \vspace*{-2cm}
    \includegraphics[scale=0.9]{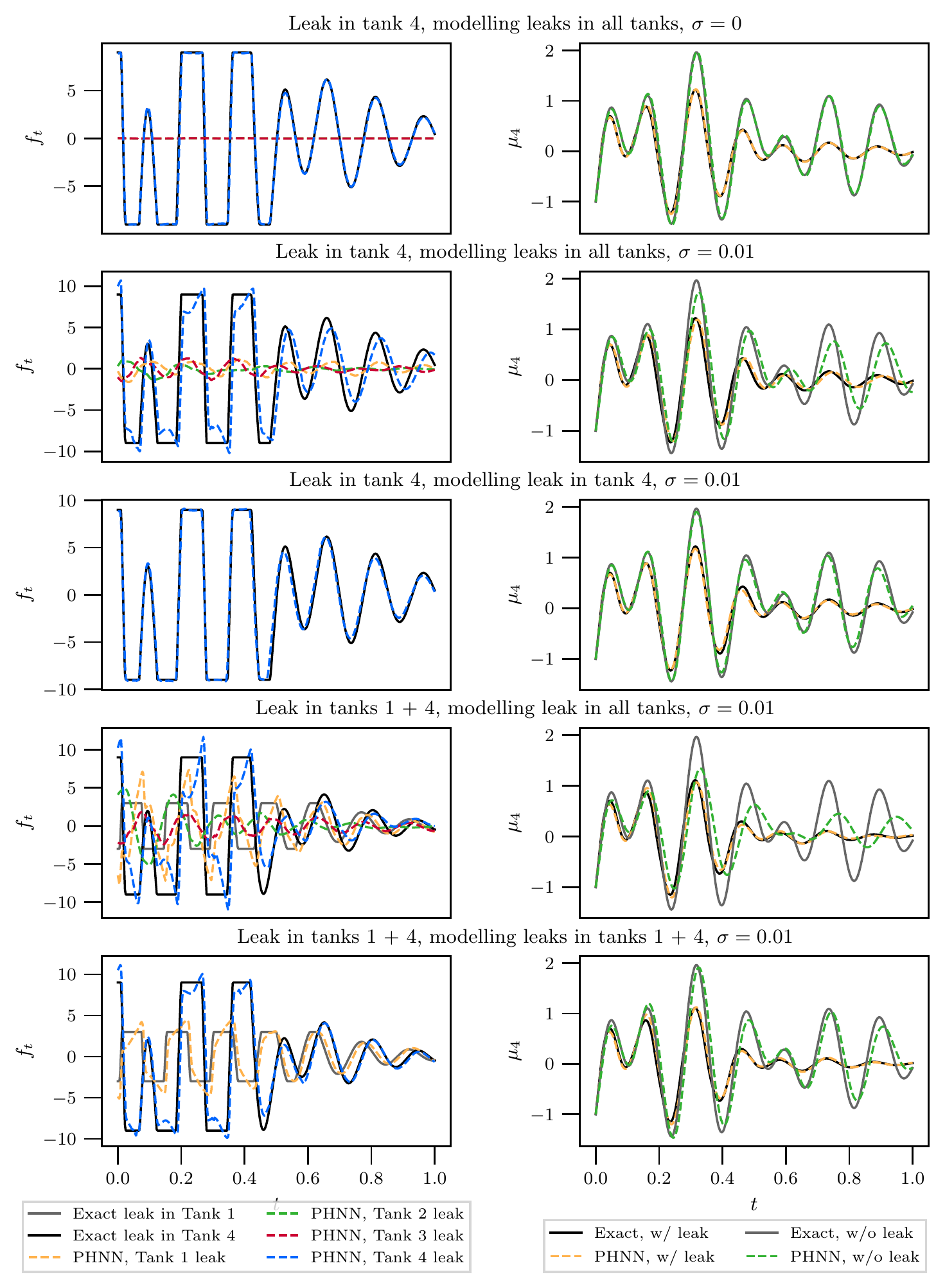}
    \caption{Left column: The external forces learned by the PHNNs trained on data gathered with one or two leakages, assuming that the location of the leakages are either known or unknown. Right column: The level in the fourth tank estimated by the PHNN before and after the leak is stopped and the PHNN external force model is set to zero. Initial condition: $\phi^0 = (-1, -1, 0, \frac{1}{2}, -1)$, $\mmu^0= (1, 1, -\frac{1}{2}, -1)$.}
\label{fig:tank_leak}
\end{figure}

Consider the example tank system described in Section \ref{subsec:tank_system}, where we now let
\begin{equation}\label{eq:leak4}
f_{t,4}(\phi,\mmu) = -30 \, \mathrm{min} (0.3, \, \mathrm{max}(\mmu_4, -0.3))
\end{equation}
model an undetected leak in the fourth tank. Training a PHNN model on data generated with this leakage allows for learning both the tank system dynamics and the leakage when the right constraints are imposed on the PHNN. In the first experiment, we assume no knowledge of which tanks might be leaking. In order to not learn spurious solutions, we apply $L_1$-regularization to the terms in $f_t$. The upmost left plot in Figure \ref{fig:tank_leak} shows how PHNN models the external force when trained using the implicit midpoint integrator for 600 epochs with $\lambda =
\{0.3, 0.1, 0.03, 0.01\}$ changing every 150th epoch, on $300$ trajectories of length $1$ with sampling time $1/400$. The predicted leakages from tanks 1-3 are negligible, while the leak from the fourth tank is accurately predicted. When re-training the model with the assumption that there is only a leak in the fourth tank we observe more efficient training; we obtain an accurate model after only 30 epochs of training with no regularization.

The second row of Figure \ref{fig:tank_leak} shows how the learned model deteriorates when training on noisy data with a lower sampling rate. The training set now consists of $1000$ training trajectories with sampling time $1/100$ and Gaussian noise with standard deviation $\sigma=0.01$. When using the fourth-order integrator \eqref{eq:srk4} and training for $2000$ epochs with $\lambda =
\{0.3, 0.1, 0.03, 0.01\}$ changing every 500th epoch we struggle to get an accurate model. However, as seen in the third row, even on noisy data the leak in the fourth tank can be learned so well that it is difficult to distinguish it from the exact solution on visual inspection, if the model is restricted to learn only the external force affecting that tank. As seen in the right column of Figure \ref{fig:tank_leak}, the trained PHNN can still be used for prediction after the leak is removed, since the external force network of the PHNN can be removed at will.

We now add an additional leak in the first tank, given by
\begin{equation*}
f_{t,1}(\phi,\mmu) = -10 \, \mathrm{min} (0.3, \, \mathrm{max}(\mmu_1, -0.3)).
\end{equation*}
Again we gather a training set of $1000$ training trajectories with sampling time $1/100$ and Gaussian noise with standard deviation $\sigma=0.01$. We follow the training procedure from the previous experiment, first avoiding to make assumptions about the location of the leaks, as shown in the fourth row of Figure \ref{fig:tank_leak}. The PHNN struggles to learn in this case, and does not generalize well to the scenario where the leaks are removed. When assuming that the leaks are in the first and fourth tanks only, results improve, as seen in the last row of Figure \ref{fig:tank_leak}.

\subsection{Control with PHNNs}\label{sec:control}
\begin{figure}[ht!]
        \centering
        \includegraphics{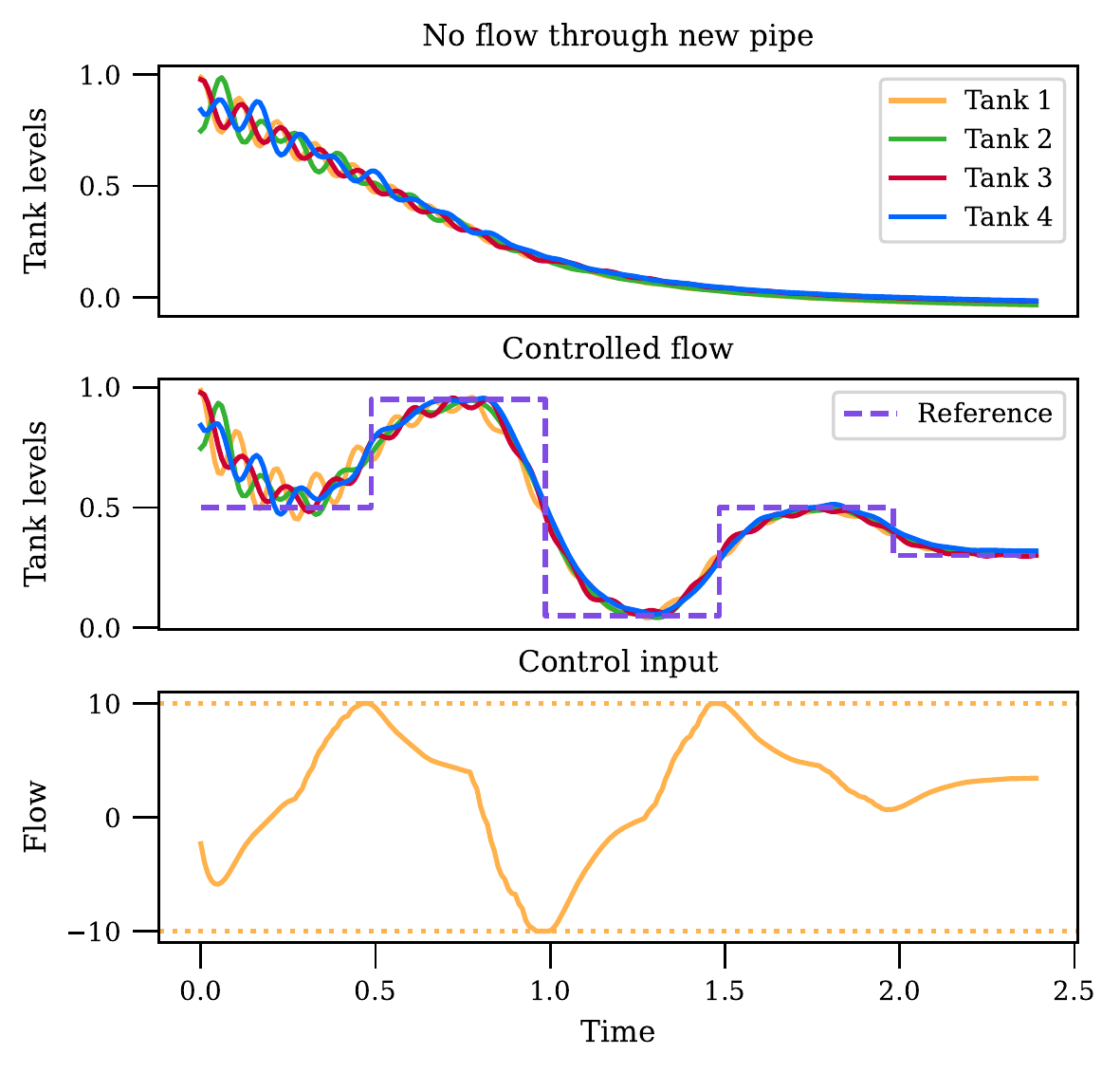}
    \caption{A learned PHNN model is used in an MPC framework, which is successfully able to drive the tank levels to the desired reference levels.}
\label{fig:control}
\end{figure}

As a last point, we highlight that the PHNN model is well suited for control, as illustrated in Figure \ref{fig:control}. In this scenario, a model is learned for the system with one leaking tank as described in Section \ref{subsec:leaks} using $1000$ data points, after which a new pipe is added to the first tank. The flow through this new pipe is controllable but also constrained with respect to minimum and maximum flow. Using the learned PHNN model in a model-predictive control (MPC) framework, the tank levels of the system can be driven to desired reference levels through the new pipe.

\section{Summary and discussion}\label{sec:discussion}
The potential advantages of a PHNN model over other data-driven models depend on the modeled system, the available data and existing knowledge of the system. As illustrated by the numerical experiments of Section \ref{subsec:msd_performance}, we may get improved performance from assuming that external forces are strictly time dependent, whether or not we impose a Hamiltonian structure on the internal system. Further, for larger systems with state-dependent external forces we are equally dependent on data quality and quantity as the baseline model, and achieve a similar prediction accuracy. However, the great advantage of using the pseudo-Hamiltonian structure in this case is the explainability and adaptability of the model; for instance, we may learn external forces and adapt to system changes, as showcased in Section \ref{subsec:leaks}. Furthermore, the integrator used during training significantly affects the resulting model, as demonstrated in Section \ref{subsec:discretization_comp}.

\subsection{Future research}\label{subsec:future_research}

The imposed pseudo-Hamiltonian structure allows for future exploitation including pseudo-Hamiltonian system identification, control with PHNN models and expansion to infinite-dimensional systems.

System identification for a more specialized formulation of what we call pseudo-Hamiltonian systems was recently proposed in \cite{Lee2021structure}, but only for the case where the external forces are known. Similarly, \cite{DiPietro2020sparse} suggest a framework for using sparse regression to obtain an analytic expression for the Hamiltonian, but only for strictly energy-preserving systems with a separable Hamiltonian. We are investigating system identification using the system set-up and the techniques presented in this paper, including the proposed symmetric fourth-order integrator \eqref{eq:srk4}, and plan to publish a paper on this in the near future.

The demonstrated benefits of the integration scheme \eqref{eq:srk4} encourages further investigation of numerical integrators tailored to the inverse problem of learning dynamical systems. Further advances in this application can offer better noise handling, for instance by taking more neighboring data points into account in the estimation of the derivative at each point. This is somewhat related to the symplectic recurrent neural networks of \cite{Chen2019symplectic}, but can be made compatible with the more general system \eqref{eq:noncangen} and a wider class of integrators.

Lastly, pseudo-Hamiltonian formulations also exist for infinite-dimensional systems \cite{Pasumarthy2004interconnections, Cardoso2019port, Brugnoli2020numerical}, and hence PHNN could be developed for finite-dimensional approximations of such systems as well.

\subsection*{Acknowledgments}
This research was supported by the industry partners Borregaard, Elkem, Eramet Norway, Norsk Hydro, Yara and the Research Council of Norway, through the projects BigDataMine (no.\ 309691) and TAPI: Towards Autonomy in Process Industries (no.\ 294544).

\bibliography{references}
\bibliographystyle{abbrv}

\end{document}